\documentclass[journal,transmag]{IEEEtran}
\ifCLASSINFOpdf
\else
\fi
\usepackage{multirow}
\usepackage{subfigure}
\usepackage{graphicx}
\usepackage{array}
\usepackage{amssymb}
\usepackage{amsmath}
\usepackage{algorithmic}
\usepackage{color}

\makeatletter

\let\NAT@parse\undefined

\makeatother

\usepackage{hyperref}  

\begin{document}
%
\title{{Frame-wise Cross-modal {Matching} for Video Moment Retrieval}}


\author{\IEEEauthorblockN{Haoyu Tang\IEEEauthorrefmark{},
		Jihua Zhu\IEEEauthorrefmark{}, 
		Meng Liu\IEEEauthorrefmark{}, \emph{Member}, \emph{IEEE},
		Zan Gao\IEEEauthorrefmark{}, and
		Zhiyong Cheng\IEEEauthorrefmark{}}
	\thanks{
		Z. Cheng is the corresponding author.	
		
		 H. Tang is with the School of Software Engineering, Xi’an Jiaotong University, Xian 710049, China, and with Shandong Artificial Intelligence Institute, Qilu University of Technology (Shandong Academy of Sciences), Jinan 250014, China. (e-mail:  tanghao258@stu.xjtu.edu.cn)
	
		J. Zhu is with the School of Software Engineering, Xi’an Jiaotong University, Xian 710049, China. (e-mail: zhujh@stu.xjtu.edu.cn)
					
		M. Liu is with Shandong Jianzhu University, Jinan 250101, China. (e-mail: mengliu.sdu@gmail.com)
					
		Z. Gao, Z. Cheng are with Shandong Artificial Intelligence Institute, Qilu University of Technology (Shandong Academy of Sciences), Jinan 250014, China. (e-mail: zangaonsh4522@gmail.com, jason.zy.cheng@gmail.com)

		This work was done when H. Tang was an intern at Shandong Artificial Intelligence Institute, Qilu University of Technology (Shandong Academy of Sciences).

}}


\markboth{Journal of \LaTeX\ Class Files,~Vol.~14, No.~8, August~2015}%
{Shell \MakeLowercase{\textit{et al.}}: Bare Demo of IEEEtran.cls for IEEE Transactions on Magnetics Journals}
%



\IEEEtitleabstractindextext{%
	\begin{abstract}
		Video moment retrieval targets at retrieving a golden moment in a video for a given natural language query. The main challenges of this task include 1) the requirement of accurately localizing (i.e., the start time and the end time of) the relevant moment in an untrimmed video stream, and 2) bridging the semantic gap between textual query and video contents. To tackle those problems, early approaches adopt the sliding window or uniform sampling to collect video clips first and then match each clip with the query to identify relevant clips. Obviously, these strategies are time-consuming and often lead to unsatisfied accuracy in localization due to the unpredictable length of the golden moment. To avoid the limitations, researchers recently attempt to directly predict the relevant moment boundaries without the requirement to generate video clips first. One mainstream approach is to generate a multimodal feature vector for the target query and video frames (e.g., concatenation) and then use a regression approach upon the multimodal feature vector for boundary detection. Although some progress has been achieved by this approach, we argue that those methods have not well captured the cross-modal interactions between the query and video frames.  	
		
		In this paper, we propose an Attentive Cross-modal Relevance Matching (ACRM) model which predicts the temporal boundaries based on an interaction modeling between two modalities. In addition, an attention module is introduced to automatically assign higher weights to query words with richer semantic cues, which are considered to be more important for finding relevant video contents. Another contribution is that we propose an additional predictor to utilize the internal frames in the model training to improve the localization accuracy. Extensive experiments on two public datasets TACoS and Charades-STA demonstrate the superiority of our method over several state-of-the-art methods. Ablation studies have been also conducted to examine the effectiveness of different modules in our ACRM model.  
	\end{abstract}
	
	}

\maketitle

\IEEEdisplaynontitleabstractindextext

%
\IEEEpeerreviewmaketitle

\section{Introduction}

\IEEEPARstart{V}{isual}-language {learning and understanding plays an important role in developing artificial intelligence in human-computer interactions} \cite{Gao_2017_ICCV,anne2017localizing, chen2019semantic,xu2019multi,zhu2020multimedia,gao2017video,yan2019stat}. Particularly, video retrieval has drawn significant attention over the past decades. Given a text query, such as ``person put a notebook in a bag", {the goal of video retrieval is to find videos that contain most relevant content with respect to the query.} To watch the specific video clip which is relevant to the query, we need to browse {the entire video to localize the relevant part in a video}, which could take hours, especially in the surveillance video scenarios. Therefore, it is important to find relevant video clips with accurate temporal boundaries (i.e., the start time and the end time) for a given query, which is the so-called video moment retrieval task. This is a recently emerged research topic and has attracted increasing attention due to its practical value \cite{anne2017localizing, chen2019semantic}.  

\begin{figure*}[htbp]
	\centering
	\includegraphics[width=0.95\textwidth]{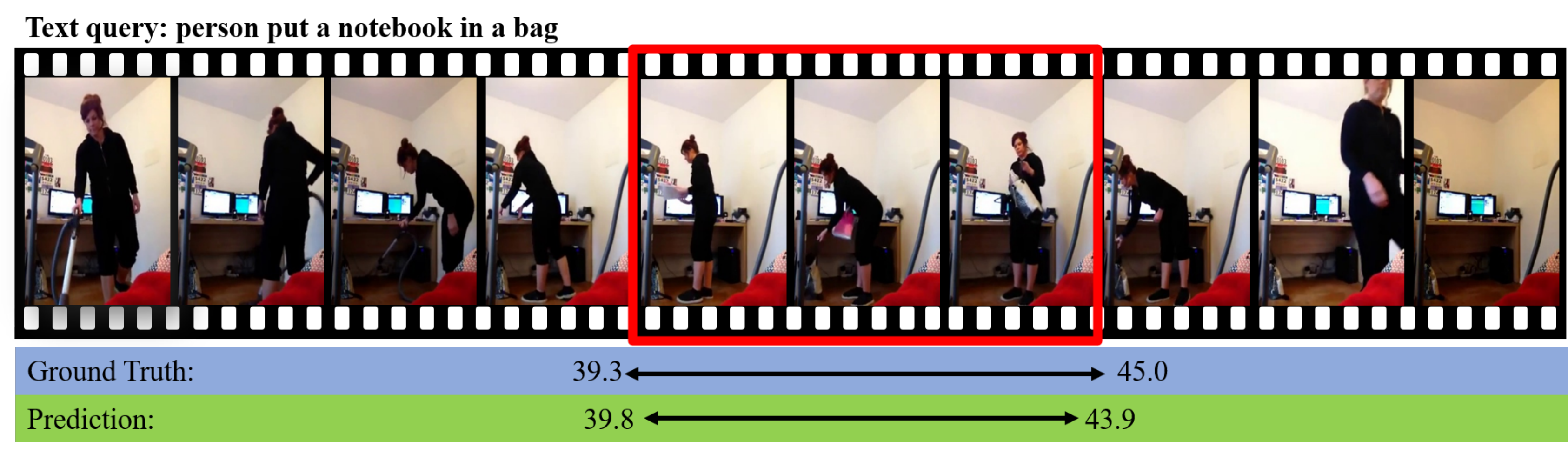}
	\caption{Moment retrieval aims to localize the temporal boundaries with a start time (39.3s) and an end time (45.0s) of a desired moment in the red box, corresponding to the given query ``Person put a notebook in a bag.".}
	\label{introduction}
\end{figure*}

Particularly, the target of moment retrieval is to precisely localize a moment in the untrimmed video whose content is in accordance with the given arbitrary natural descriptions \cite{Gao_2017_ICCV}, as illustrated in Figure \ref{introduction}. Based on the experience of approaches in video retrieval, early approaches follow a two-step manner, i.e., {first generating the moment candidates via the temporal sliding window strategy and then matching them with the query in a common cross-modal space} \cite{Gao_2017_ICCV, liu2018cross, liu2018attentive, ge2019mac, jiang2019cross}. Because the desired moments can be of varying lengths, various sizes of sliding windows need to be employed to generate numerous overlapping segments to match with the query. Therefore, this type of method is cumbersome and resource-consuming. 

To reduce the number of moment candidates, several methods have been developed, such as uniformly sampling segments in the video \cite{chen2018temporally}, {while others leveraging the segment proposal network \cite{xu2017r} to generate moment candidates that most likely contain the potential contents} \cite{ xu2019multilevel}. However, those methods still need to generate moment candidates and then match them with the query, resulting in inferior efficiency. Moreover, although these approaches refine the boundaries of the selected moment candidates, the performance will still be far from satisfactory when the selected candidate has a small overlap with the ground-truth moment.

To overcome these drawbacks, several one-step retrieval methods without the requirement of generating video segments have been proposed~\cite{hahn2019tripping,wu2020tree}. Specifically, the cross-modal feature of the query and each video frame is used to predict the probability of the frame to be the starting or ending frame of the target moment. The concatenation of the query feature and frame feature is often used in the previous method to generate the cross-modal feature~\cite{ghosh-etal-2019-excl,chen2020rethinking,zhang-etal-2020-span,yuan2019find}. Although significant improvement has been achieved, there are still several limitations. { First, in those methods, the video features {or} query features are extracted via separate networks. As a result, the video feature and query feature are from different feature spaces, but they are directly concatenated for the next step prediction. In addition, the concatenation of these features cannot well capture the interactions between the query and video content. Besides concatenation, other methods use the query feature to attend the video frame to generate a weighted video feature for boundary prediction~\cite{yuan2019find, Rodriguez_2020_WACV}}. Similarly, they fail to model the fine-grained interactions between video frames and query words, which are important to guide the localization. Besides, the problem of different feature spaces has not been tackled in those methods either. {Intuitively, the relevant video frames should be closer to the query than the irrelevant ones in the shared feature space,  which is a common assumption in the cross-modal retrieval problem. From this perspective, the concatenation or the attended video features in the above methods cannot well model the interaction between the frame feature and query feature.}
Furthermore, most existing methods mainly focus on {directly} predicting the temporal boundaries and overlook the internal frames (i.e., the ones between the start and end frames), which also contain valuable information~\cite{zhang-etal-2020-span}. For the query “person put a notebook in a bag” as illustrated in Figure.\ref{introduction}, the desired moment in the video is the process of the person picking up the book and putting it into the bag. Notably, as the camera is locked in a specific view, the notebook has not even appeared outside the desired moment and thus the semantic information in all frames of the desired moment is similar. Under such circumstances, {the cross-modal features extracted by the model on these frames within the moment become very similar}, resulting in indistinguishable prediction scores between the internal frames and the boundary frames, which is not beneficial for the judgment of the boundaries. If the model also predicts whether a frame belongs to the desired moment, the information of internal frames can be leveraged to facilitate the boundary prediction. 

Based on the above considerations, we propose a novel moment retrieval model called Attentive Cross-modal Relevance Matching (ACRM), which directly predicts the relevant moment boundaries without the requirement of pre-processed candidates of video clips. Specifically, we process each frame of a target video and embed the frame feature into the same space with the query feature. {Rather than employing the concatenation or the attended video features, we use an interaction function to model the interactions between the frame and the query for boundary prediction.} In addition, a multimodal attention module is introduced to estimate the importance of each word in the query. This can not only enhance {the cross-modal matching} between video frames and queries but also exploit the fine-grained frame-query interactions. Besides, we incorporate the prediction of internal frames as element moment frames into the objective function, which can effectively improve the boundary prediction accuracy. Extensive experiments have been conducted on two benchmark datasets Charades-STA \cite{Gao_2017_ICCV} and TACoS \cite{rohrbach2012script}. The results show that our model can consistently outperform all the competitors by a large margin. 

In summary, the main contributions of this work are three-fold:

\begin{itemize}
	\vspace{-2pt}
	\item We highlight the importance of modeling the interactions between the cross-modal features for video moment retrieval and propose a novel ACRM model. In particular, an attention mechanism and {a similarity function are integrated into ACRM to model the interactions between the video frame and query features.}
	\item To fully leverage the information containing in the internal frames of the moment, we add an extra predictor to estimate the probability of a frame to be the internal frame, which is proven to be effective in our experiments.
	\item We conducted extensive experiments to evaluate the performance of our proposed model by comparing it to several state-of-the-art methods. We also analyze the effectiveness of each module in our model by ablation studies. Our code has been released for the reproduction of the experiments.\footnote{https://github.com/tanghaoyu258/ACRM-for-moment-retrieval}
\end{itemize}

\section{Related Work}
\subsection{Moment Retrieval} 
Finding the desired moment in an untrimmed video according to a sentence query is a challenging task due to the requirement of cross-modal understanding. To accomplish this task, early methods follow a two-step manner, which generates moment candidates first and then matches them with the query~\cite{anne2017localizing} {to find the most relevant one}. For example, Gao et al. \cite{Gao_2017_ICCV} presented a Cross-modal Temporal Regression Localizer (CTRL), which generates moment candidates via a temporal sliding window method, and then encodes those candidates and the sentence query into the same space to find the relevant candidates by matching the candidates with the query in the space. Following this framework, Liu et al. proposed a ROLE model \cite{liu2018cross} which uses a language temporal attention module to learn sentence representation, and an ACRN model \cite{liu2018attentive} which adopts a memory attention network to capture the contextual information of the moments. Although {these methods achieve better performance} with advanced representation learning methods, they are still resource-consuming due to the sliding window strategy.

To overcome this limitation, some research efforts have been dedicated to {reducing the number of the temporal moment candidates~\cite{chen2018temporally,zhang2019man}}. For instance, inspired by the R-C3D \cite{xu2017r} model which was designed for action localization in the video, Xu et al. \cite{xu2019multilevel} employed a segment proposal network \cite{xu2017r} to generate the varied-length moment candidates, and presented a multilevel language and vision integration model which incorporates sentence features to generate the attended moment candidates. Wang et al. \cite{wang2020temporally} {proposed a temporal grounding model that explores the interactions between video sequence and sentence with the Match-LSTM \cite{wang2016learning} to score multiple candidates at each time step}. Zhang et al. \cite{zhang2020learning} employed a downsampling strategy to reduce the number of candidates obtained from a two-dimensional temporal map.

Recently, researchers attempt to directly localize the desired moments without the requirement of generating candidate first. Several methods have been proposed in this direction. For instance, the reinforcement learning (RL) strategy has been adopted in moment retrieval~\cite{hahn2019tripping,wu2020tree}. In general, the RL-based methods progressively update the temporal boundaries over the entire video to locate the desired moment for a given query. Meanwhile, another research line is to predict the probabilities of each frame to be the boundary frame based on the cross-modal feature of the query and video frame. {For example, Yuan et al.~\cite{yuan2019find} proposed an ABLR model that concatenates the attended query feature and the video frame features and then regresses the temporal interval to the boundaries for each frame}. The ExCL model~\cite{ghosh-etal-2019-excl} used a regression method upon the concatenation of the query and video frame feature to predict the boundary frame. 
Following this work, Rodriguez et al. \cite{Rodriguez_2020_WACV} predicted the boundary based on the attentive video features, which are attended by the query feature. Chen et al. \cite{chen2020rethinking} and Zhang et al. \cite{zhang-etal-2020-span} concatenated the query feature, video feature, and their similarity together for the subsequent prediction of the temporal boundaries. We argue that the above methods (i.e., concatenation of query and video frame feature, and the attended video features) cannot well capture interactions between cross-modal features. In this work, we encode the query and video frame into the same feature space to obtain a similarity vector, upon which a prediction model is used to detect the boundary. 

{In addition, we adopt an attention mechanism to identify meaningful query words and add an internal frame predictor into the objective function to enhance the boundary detection. Some previous methods have also adopted attention mechanisms, such as \cite{chen2018temporally} and \cite{wang2020temporally} employing the Match-LSTM \cite{wang2016learning} attention. However, those methods still score multiple moment candidates and concatenate the query and video features, while our method directly predicts the boundary and uses the interaction function to model the cross-modal interactions.} There are also methods exploiting the information of the internal frames. Rodriguez et al.~\cite{Rodriguez_2020_WACV} proposed to highlight the temporal attention weights across the internal frames, and Zhang et al.~\cite{zhang-etal-2020-span} temporally extended the region of the internal frames to include more video context. Different from those methods by extending the region of the internal frames, our model exploits the information of internal frames by using an additional predictor in the objective function. 

\subsection{Temporal Action Localization} 
{The temporal action localization task also needs to localize the boundary of required moments in videos.} The difference is that they have a pre-defined action list and the required moments are action instances in the list. As this task has pre-defined concepts, most methods are supervised. A general pipeline is to first generate candidate clips containing activities and then use the pre-trained action classifiers to detect the action. For instance, Gao et al. \cite{gao2017turn} jointly classified the action proposals and fine-tuned their temporal boundaries in a temporal unit regression network, and Xu et al. \cite{xu2017r} introduced a Region Convolutional 3D model (R-C3D) to generate temporal candidates containing activities {for the final classification}. Ma et al.~\cite{ma2016learning} proposed to incorporate the predicted scores from a temporal LSTM over the detection span. There are also approaches designed in a weakly-supervised fashion. Wang et al. \cite{wang2017untrimmednets} integrated the classification module and the selection module to predict the action proposals and then select the most probable ones, respectively. Paul et al. \cite{paul2018w} leveraged an attention-based module along with multiple instances learning to learn pairwise video similarity constraints for localization and classification. {Shou et al. \cite{shou2018autoloc} employed an IOC loss to minimize the distance between the frame features of the action and the features of the context frames outside the action. Lin et al. \cite{lin2018bsn} proposed a boundary sensitive network (BSN) model to leverage the information of the internal frames for this task. It is worth mentioning that our method directly maximizes the averaged log-likelihood of the internal frames of the ground-truth moment, while BSN \cite{lin2018bsn} generates some candidate segments and then feeds randomly sampled N frame features to maximize the log-likelihood of these sampled frames from every candidate segment.}

Despite great progress has been achieved for the task of temporal action localization, those methods are not suitable for the moment retrieval task, because we do not have knowledge about the queries. {In fact, video moment retrieval is a more complex task, as the query can be of arbitrary lengths and a variety of concepts.}


\begin{figure*}[htbp]
	\centering
	\includegraphics[width=0.95\textwidth]{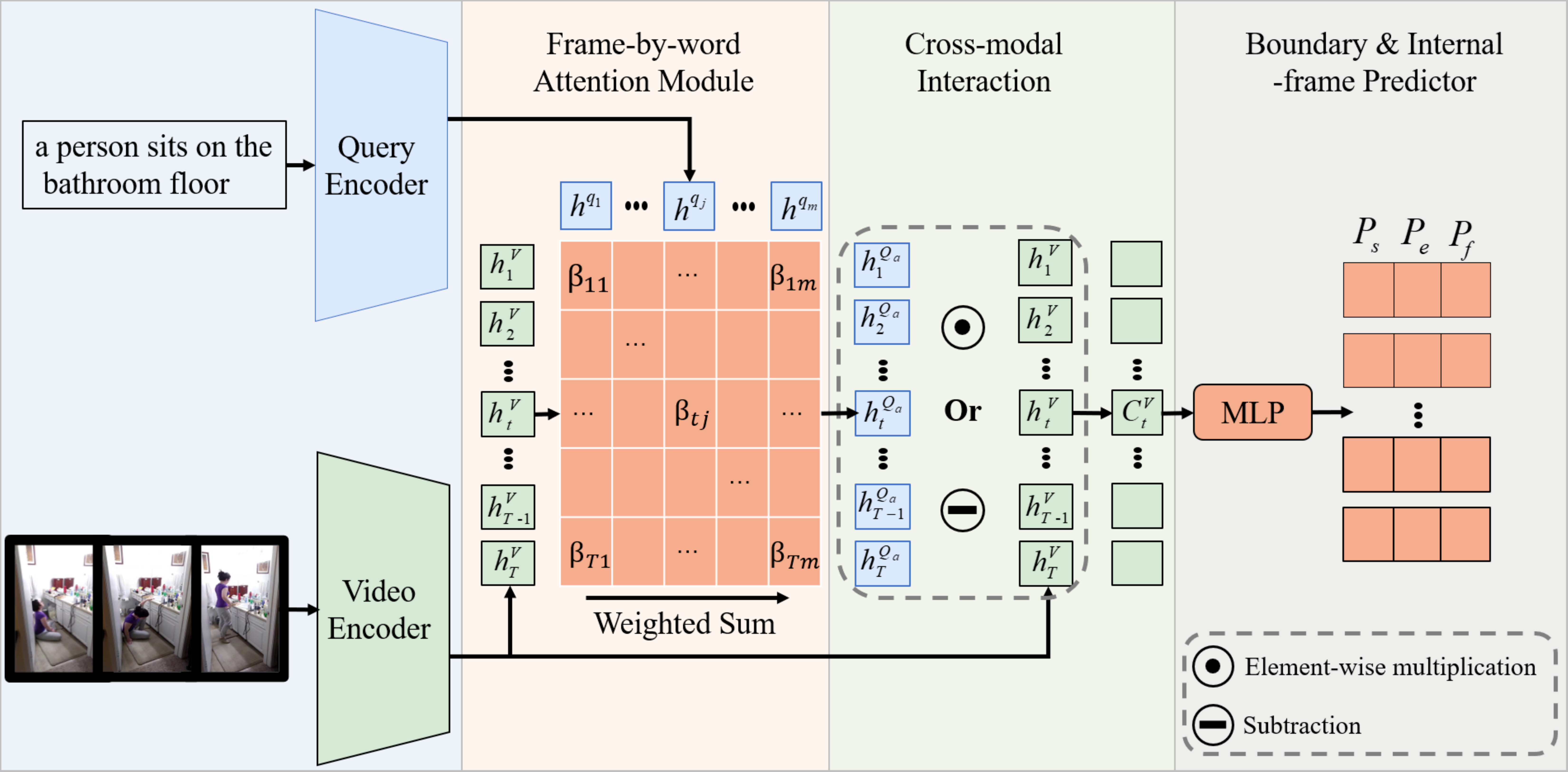}
	\caption{{An illustration of the detailed ACRM model. It comprises of four components: two encoders to extract the video frame features and textual embeddings separately; an attention module to generate frame-specific query representation; a cross-modal interaction function to calculate the cross-modal interactions; and a two-branch predictor to estimate the temporal boundaries.}}
	\label{fig:model}
\end{figure*}

\subsection{Sentence-based Video Retrieval} 
Another closely related task is sentence-based video retrieval, which aims to search the most relevant video from a video set. Many methods regard it as a ranking problem by mapping the videos and sentences into a common space \cite{feng2017computational,feng2018learning,bojanowski2015weakly,lin2014visual,dong2018predicting}. A general approach is to feed the video into a pre-trained CNN model to extract frame features, which are aggregated into a video feature by mean-pooling. Such processing leads to inefficiency in learning a common embedding space since the temporal cues cannot be captured. To tackle this problem, Dong et al.~\cite{dong2019dual} proposed to encode global and local pattern information for both video and text ends. Lin et al.~\cite{lin2014visual} presented to retrieve the video by matching a semantic graph from parsed sentence descriptions and the visual concepts in the video.  Besides, Mithun et al.~\cite{mithun2018learning} proposed to learn two different embedding spaces to obtain temporal and appearance information. {The advancement of sentence-based video retrieval is beneficial to the development of video moment retrieval task, especially the techniques of matching query and video contents}, as both tasks need to find relevant video content for the targeted query. The difference is that video moment retrieval needs to localize the boundary of the relevant video clips, which is more challenging than just find the relevant videos from a video set. 


\section{The Proposed Model}
\subsection{Preliminaries} 
Before formally describing our model in detail, {we first introduce the primary notations}. Given a video $V=\{{{v}_{t}}\}_{t=1}^{T}$, where $v_t$ represents the image frame at time $t$ and $T$ is its length, and a sentence query $Q=\{ {{q}_{j}} \}_{j=1}^{m}$ with $m$ words, our goal is to find {the most relevant moment in the video} by identifying the accurate start and end frames. Formally, this problem can be formulated as a mapping function: 
\begin{equation}
{{L}_{\theta}}:(V,Q)\to({{t}^{s}},{{t}^{e}}),{{t}^{s}}<{{t}^{e}} 
\end{equation}
where $t^s$ and $t^e$ represent the start time and the end time of a golden video moment, respectively. The proposed model is trained in an end-to-end fashion on the training set, which contains $K$ instances. Each instance is a video-query-boundaries tuple $\{{V},{{Q}},\tau^{s},\tau^{e}\}$, where the query ${Q}$ is associated with the ground-truth start and end time point $\tau^{s}$ and $\tau^{e}$ in the video $V$. During the evaluation stage, given an unseen video-query pair $\{{V}_{e},{Q}_{e}\}$, the goal is to predict the temporal boundaries $\{{t}^{s'},{t}^{e'}\}$ in the video ${V}_{e}$.
\subsection{Our Model}
Figure \ref{fig:model} shows an overview of our model, which consists of four components: 1) \textbf{Feature extraction} module extracts the video frame and query features by two separate networks; 2) \textbf{Cross-modal interaction} module models the interaction between the features of two modalities; 3) \textbf{Attention Module} attentively fuses the cross-modal features to generate the frame-specific query representation { to further enhance the interaction}; and 4) \textbf{Prediction module} estimates the probability of a frame to be a boundary frame and an internal frame. In the next, we introduce the four modules in sequence.

\subsubsection{Feature Extraction}
In our model, the features of video frames and queries are extracted by two separate networks. 

\textbf{Video feature extraction.} We apply the off-the-shelf visual feature extractors to extract the features of each frame for an untrimmed video $V=\{{{v}_{1}},\dots, v_T\}$, such as the I3D~\cite{carreira2017quo} or the C3D extractor\cite{tran2015learning}. The {BiLSTM~\cite{schuster1997bidirectional} is employed here} to sequentially process the extracted visual feature because it encodes video sequence from bi-directions. In this way, every frame representation will be affected by its contiguous frames spontaneously to augment the incorporated contextual information. {Specifically, every hidden state of the forward and the backward LSTM are concatenated together to obtain the new representations. The video encoder is defined as follows:
\begin{equation}
\label{equ:visualextractor}
\begin{aligned}
& {\mathbf{f}_{t}}=F\left( {{v}_{t}} \right)\\ 
& \overrightarrow{\mathbf{h}_{t}^{V}}=\overrightarrow{\text{LSTM}}({{\mathbf{f}}_{t}},\overrightarrow{\mathbf{h}_{t-1}^{V}})\\
&\overleftarrow{\mathbf{h}_{t}^{V}}=\overleftarrow{\text{LSTM}}({{\mathbf{f}}_{t}},\overleftarrow{\mathbf{h}_{t\text{+}1}^{V}}) \\ 
\end{aligned}
\end{equation}
where $F(\cdot)$ represents the visual extractor, and $\mathbf{f}_t$ is the extracted feature of the $t$-th frame. $\mathbf{h}^{V} = [\overrightarrow{\mathbf{h}_{t}^{V}};\overleftarrow{\mathbf{h}_{t}^{V}}] \in {{R}^{T\times {{d}}}}$ is the concatenated video features of two LSTM and $d$ is the dimension of the feature vector}.

\textbf{Query feature extraction.}
As for the sentence query $Q$, the pre-trained glove~\cite{pennington2014glove} embedding is used to transform the query words $Q=\{{{q}_{1}},\dots, {{q}_{m}}\}$ into embeddings $\mathbf{S}=\{{\mathbf{s}_{1}},{\mathbf{s}_{2}},\cdot \cdot \cdot ,{\mathbf{s}_{m}}\}$, i.e. $\mathbf{S}=glove({Q})$. Note that here we can use any other word embedding approaches such as Skip-Thought~\cite{kiros2015skip}. With the sequential word embeddings $S$ as input, {a BiLSTM network} is used to encode the sentence into the representation ${\mathbf{h}^{Q}}=\{{\mathbf{h}^{{{q}_{1}}}}\text{,}{\mathbf{h}^{{{q}_{2}}}}\text{,}\cdot \cdot \cdot \text{,}{\mathbf{h}^{{{q}_{m}}}}\}\in {{R}^{m\times {d}}}$, where ${d}$ is the dimension of the hidden state, which is the same as the feature vector of the video frame. Compared to glove methods, {the BiLSTM} could comprehensively encode the context information of the whole sentence.  

\subsubsection{Cross-modal Interaction}
Previous methods like~\cite{ghosh-etal-2019-excl,liu2018cross} directly concatenate the extracted video frame feature and query feature for the next step prediction. However, the frame feature and query feature are extracted by different networks. {On one hand, the learned features based on two networks are from different spaces, the direct concatenation is problematic. On the other hand, the values of frame feature and that of query features might be in different range. As a result, the concatenated feature will be of great variance, which increases the difficulty of learning a good prediction model in the next step. Besides, the concatenation cannot well capture the interaction between the two modalities. In cross-modal retrieval, a successful approach for the cross-modal match is to compute similarity (such as element-wise
multiplication or Euclidean distance) between the feature vectors of two modalities in the same feature space.} Motivated by this observation, in this work, we propose to use different interaction functions to model the interaction between the frame and query features.

Specifically, we first embed the video frame feature and the query feature into a common space, in which the interactions between the frame and query are modeled. Formally, the video and query features are embedding to the same space by a transformation matrix:
\begin{equation}
\label{equ:embednorm}
\left\{ \begin{aligned}
& {{{\mathbf{\hat{h}}}}^{Q}}=\mathcal{N}\left( \mathbf{W_q}{\mathbf{h}^{{Q}_{a}}}+\mathbf{b_q}\right) \\ 
& {{{\mathbf{\hat{h}}}}^{V}}=\mathcal{N}\left( \mathbf{W_v}{\mathbf{h}^{{V}}}+\mathbf{b_v}\right) \\ 
\end{aligned} \right.
\end{equation}
{where $\mathbf{W_q}$ and $\mathbf{W_v}$ are trainable weight matrices, and $\mathbf{b_q}$ and $\mathbf{b_v}$ are bias vectors. $\mathbf{h}^{{Q}_{a}}\in {{R}^{T\times {{d}}}}$ is the query feature calculated through the attention module, which we will explain in the section~\ref{sec:att}. $\mathbf{h}^{{V}}\in {{R}^{T\times {{d}}}}$ is the extracted video feature from Eq.\ref{equ:visualextractor}. $\mathcal{N}(\cdot)$ is an operation to transform the values of the query and video frame features into the same range (e.g., [-1, 1]).} Different strategies can be used here to achieve the goal. Here, we test two popular approaches. One is to use an activation function which are widely used in neural networks and the $\textbf{tanh}$ activation function is used here; the other one is the normalization method and the Gauss distribution normalization is adopted. 
Finally, the interaction between the frame and query are modeled: 
\begin{equation}
\label{equ:crossfeat} 
\mathbf{C^V}=f(\mathbf{\hat{h}}^V,\mathbf{\hat{h}}^{{Q}_{a}}) \\ 
\end{equation}
where $f(\cdot)$ is the interaction function. Two simple yet effective interaction functions are explored in this work: \textbf{element-wise multiplication} and \textbf{subtraction}.

\subsubsection{Attention Module}
\label{sec:att}
Since the localization of moment boundaries needs the frame feature and query feature, the integration of obtained word embeddings into a discriminative representation of the query is crucial.
{A widely-used strategy} is the mean pooling of the embedding from all the hidden states. A limitation of this method is that it treats {each word} in the query equally. However, it is common that some words convey more information to localize the relevant frames. For example, for the query ``the black car is arriving", the word ``arriving" conveys crucial temporal information, which is thus more helpful on identifying the desired moments. {The mean pooling cannot distinguish the different importance of words, and thus fails to identify the relevant moments in such video}. 


To tackle this issue, we design an attention module to generate frame-specific query representation to {further explore the relations between the query and the video content}. Specifically, it employs the $t$-th frame feature $\mathbf{h}_{t}^{V}$ to adaptively attend all word features in the query $\mathbf{h}^{Q}\in {{R}^{m\times {d}}}$ to obtain a summarized query feature $\mathbf{h}_{t}^{{Q}_{a}}$. Accordingly, the attention weight represents the relevance between the $t$-th frame and each words in the query. The attention module is detailed as:
\begin{equation}
{{r}_{tj}=\mathbf{w}_{r}^{T} \cdot \tanh \left(\mathbf{W}_{s} \mathbf{h}^{{q_j}}+\mathbf{W}_{v} \mathbf{h}_{t}^{V}+\mathbf{b}_{r}\right)} 
\end{equation}
\begin{equation}
\label{Eq:softmax}
{{\beta}_{t j}=\frac{\exp \left({r}_{t j}\right)}{\sum\nolimits_{k=1}^{m} \exp \left({r}_{t k}\right)}} 
\end{equation}
where the weight matrix $\mathbf{W}_{s}$ and $\mathbf{W}_{v}$ encode the hidden state of the $t$-th frame and the hidden state of the $j$-th word into a common space to compute ${r}_{tj}$, which is fed in Eq.\ref{Eq:softmax} to obtain the normalized attention weight ${\beta}_{t j}$. $\mathbf{w}_{r}^{T}$ is a trainable vector. The $t$-th query feature is summarized as:

\begin{equation}
{\mathbf{h}_{t}^{{Q}_{a}}=\sum\nolimits_{j=1}^{m}\left({\beta}_{t j} \cdot \mathbf{h}^{{q}_{j}}\right)} 
\end{equation}
After each frame is processed, the obtained frame-specific query features are concatenated as $\mathbf{h}^{{Q}_{a}}\in {{R}^{T\times {{d}}}}$, which is the frame-specific query feature in Eq.~\ref{equ:embednorm}. 

\subsubsection{Prediction}
To estimate the start and end frame of a specific moment, the prediction module processes the obtained cross-modal features and outputs the prediction vector with the same length $T$ as the input frames. In previous methods, the predictors only consider the prediction of the start frame and end frame by maximizing the scores of the ground-truth boundaries. They have not exploited the information of internal frames between the boundaries in the model training. In fact, those internal frames contain rich information to help the localization of the temporal boundary. For example, if an internal frame matches the query well, it should be in the desired moment video clips.

Based on the above considerations, we integrate an additional internal frame predictor to estimate the probability of a frame to be the internal frame. The prediction of a frame to be a start frame, end frame, or the internal frame of the desired moment follows the same pipeline with separate prediction networks. Many prediction functions can be used here, such as the Multi-layer Perceptron (MLP) with a regression loss or an LSTM with a classification loss. The tied LSTM predictor with the classification loss is selected as the backbone of our prediction module because of its simplicity. In the next, we take the prediction of the start frame as an example to describe the process. Specifically, for the frame sequence,  the cross-modal features (obtained by Eq.~\ref{equ:crossfeat}) are first processed by an LSTM, whose outputs are then fed into an MLP network. Finally, the softmax function is used to predict its probability to be the start frame. {Formally, given the cross feature of the frame-query pair $\mathbf{C^V}$,                                                                                                                                                                                                                                                                                                                                                                                                                                                                                                                                                                                                                                                                                                                                                               
\begin{equation}
\mathbf{h}_{t}^{P}=\text{BiLSTM}({\mathbf{C}^{V}_{t}},\mathbf{h}_{t-1}^{P})
\end{equation}
where $\mathbf{h}_{t}^{P}$ is the $t$-th hidden state of the BiLSTM. All the obtained hidden states are concatenated as $\mathbf{h}^{P} \in{{R}^{T\times{d}}}$}, which is fed in the MLP layer as follows:
\begin{equation}
\mathbf{e}_{s}=MLP\left( \mathbf{h}^{P}\right)\\
\end{equation}
\begin{equation}
\mathbf{P}_{s}=softmax{\left(\mathbf{e}_s\right)}
\end{equation}
where $\mathbf{e}_{s}$ is the output vector of MLP and $\mathbf{P}_{s}\in{{R}^{T}}$ is the start frame probabilities vector. 


\subsection{{Loss Function}}
The objective function consists of two parts. Firstly, a classification loss, which encourages the correct start and end frame to have a larger prediction score by maximizing the log-likelihood of the ground-truth boundaries, is employed for boundary detection. Secondly, {for each video-query pair, we would like the frames inside the moments to have a larger prediction score than the outside ones.} This is achieved by maximizing the averaged log-likelihood of the internal frames.
\begin{equation}
{{L}_{c}}=-\frac{1}{K}\sum\limits_{i=1}^{K}{\log \left( {\mathbf{P}_{s}}\left( \tau _{i}^{s} \right) \right)+\log \left( {\mathbf{P}_{e}}\left( \tau _{i}^{e} \right) \right)} 
\end{equation}
\begin{equation}
\label{equ:internal_loss}
{{L}_{I}}\text{=}-\frac{1}{K}\sum\limits_{i=1}^{K}{\sum\limits_{j=s}^{e}{\log \left( {\mathbf{P}_{f}}\left( \tau _{i}^{j} \right) \right)}/\left( \tau _{i}^{e}-\tau _{i}^{s} \right)}
\end{equation}
where $\mathbf{P}_{s}$, $\mathbf{P}_{e}$, $\mathbf{P}_{f}$ is the start frame, end frame, and the internal-frame probabilities, respectively. $\tau_{i}^{s}$ and $\tau_{i}^{e}$ represents the ground-truth temporal boundaries of the $i$-th video-query pair. The overall loss function is summarized as:
\begin{equation}
\label{loss function}
L\text{=}{{L}_{c}}+\lambda{{L}_{I}} 
\end{equation}
where $\lambda$ balances the two losses. Note that the internal-frame predictor is used to improve the learning process in our model, {and it is only used in the training stage.}

In the test stage, the boundaries are determined only by the boundary prediction as follows:
\begin{equation}
\begin{aligned}
& {{t}^{s}},{{t}^{e}}=\underset{{{t}^{s}},{{t}^{e}}}{\mathop{\arg \max }}\,{\mathbf{P}_{s}}\left( {{t}^{s}} \right){\mathbf{P}_{e}}\left( {{t}^{e}} \right) \\ 
& =\underset{{{t}^{s}},{{t}^{e}}}{\mathop{\arg \max }}\,\mathbf{e}_{s}\left( {{t}^{s}} \right)+\mathbf{e}_{e}\left( {{t}^{e}} \right)\\ 
& s.t.\,{{t}^{s}}\le {{t}^{e}} 
\end{aligned}
\end{equation}

\section{Experiment}

\subsection{Experiment Setup}
\subsubsection{Dataset} 
\textbf{Charades-STA} \cite{Gao_2017_ICCV}: This dataset was extended from the original Charades dataset in \cite{10.1007/978-3-319-46448-0_31}, which is mainly used for temporal activity localization task and only contains the video-level description. To accommodate the moment retrieval task, Gao et al. \cite{Gao_2017_ICCV} generated the sentence-clip annotations by decomposing the provided descriptions into shorter parts, which were assigned to the clips and further manually verified by annotators. We follow the experimental setting defined in~\cite{Gao_2017_ICCV}, in which the training set has 12408 video-sentence pairs and the testing set has 3720 video-sentence pairs. The videos are 30 seconds long on average.


\textbf{TACoS} \cite{rohrbach2012script}: This dataset is collected from the MPII Cooking Composite Activities dataset~\cite{rohrbach2012script}, which contains 127 long videos in the cooking scenarios. Following the dataset split as in Gao et al~\cite{Gao_2017_ICCV}, 10,146, 4,589, and 4,083 clip-sentence pairs are employed for training, validation, and testing, respectively. In the dataset, the timespans of the labeled moments are short and the overlapping between the moments is often large, which makes it a challenging dataset. 

\subsubsection{Implementation Details}
Each sentence is tokenized by Stanford CoreNLP~\cite{manning-EtAl:2014:P14-5} and then the pre-trained 300-dimensions glove embeddings~\cite{pennington2014glove} are adopted to obtain the word-level representations. {The vocabulary size is set as 3720 and 1438 for Charades-STA and TACoS, respectively. For the video frame representations, the I3D and C3D features~\cite{carreira2017quo} were extracted for both the Charades-STA dataset and the TACoS dataset. Note that the parameters of the I3D or C3D extractor and the glove embeddings are fixed during the training. The size of the visual and query bidirectional LSTM encoders are 256 dimensions, and the MLP predictor consists of a single 256-dimension hidden layer with $\textbf{tanh}$ as the activation function.} We train our model with a batch size of 64 and adopt the early-stopping strategy for both two datasets. The Adam optimizer is adopted with learning rate of 0.001. Besides, $\lambda$ is empirically set as 0.7, 1.1 for the Charades-STA and TACoS datasets, respectively. Dropout is adopted in LSTMs to prevent over-fitting and the dropout ratio is set to 0.5.

\subsubsection{Evaluation Metrics}
The standard evaluation metrics ``R@$n$, IoU=$m$" \cite{hu2016natural} and ``mIoU" \cite{ge2019mac} are used for evaluation. To be specific, the Intersection over Union (IoU) between the predicted and ground-truth temporal boundaries of the top-$n$ retrieval results for each query is computed, and ``R@$n$, IoU=$m$" is the percentage of instances that have at least one in top-$n$ retrieval results with IoU larger than the threshold ${m}$. ``mIoU" represents the mean IoU of top-$\mathbf{1}$ result across all queries.
\begin{table}
	\caption{Performance comparison between the proposed model and the state-of-the-arts on Charades-STA dataset. The symbol '*' means the I3D features are adopted.}
	\label{table:results on Charades}
	\centering
	\begin{tabular}{c|c|c|c|c}
		\hline
		\multirow{2}{*}{Method} & R@1 & R@1 & R@1 &\multirow{2}{*}{mIoU}\\ 
		& IoU=0.3 & IoU=0.5 & IoU=0.7 & \\
		\hline\hline
		CTRL(ICCV'2017) &-& 23.63 & 8.89 & -\\
		SM-RL(CVPR'2019) & - &24.36&11.17&-\\
		ABLR(AAAI'2019) &-&24.36&9.01&-\\
		SAP(AAAI'2019) &-&27.42&13.36&-\\
		ACL(WACV'2019) & - & 30.48 & 12.2 & 33.84\\
		MLVI(AAAI'2019) & 54.70 & 35.60 & 15.80 &-\\
		TripNet(CVPR'2019) & 51.33 & 36.61& 14.50&-\\
		CBP(AAAI'2020)&-&36.80&18.87&35.74\\
		GDP(AAAI'2020)& 54.54&39.47&18.49&-\\
		2D-TAN(AAAI'2020)&-&39.81&23.31&-\\
		{ACRM\textsubscript{dt}} &{59.92}&{40.78}&{22.28}&{40.48}\\
		\hline
		SCDM*(NIPS'2019)&-&\underline{54.44}&33.43&-\\
		ExCL*(EMNLP'2019) &65.10&44.10&22.60&-\\
		DRN*(CVPR'2020)&-&53.09&31.75&-\\
		VSLNet*(ACL'2020)&\underline{70.46}&54.19&\underline{35.22}&\underline{50.02}\\		
		\hline\hline
		{ACRM\textsubscript{sg}*} &69.89&50.46&31.13&48.45\\
		{ACRM\textsubscript{st}*} &72.07&56.91&36.56&51.39\\
		{ACRM\textsubscript{dg}*} &72.15&\bf{57.93}&37.15&52.60\\
		{ACRM\textsubscript{dt}*} &\bf{73.47}&57.53&\bf{38.33}&\bf{53.01}\\
		\hline
	\end{tabular}
\end{table}

\subsection{Comparison with State-of-the-Arts}
\subsubsection{Baselines} The proposed model is compared with several state-of-the-art methods.	 For a fair comparison, we directly copy the reported results from the original papers of those methods. The considered competitors are listed as follows:
\begin{itemize}
	\vspace{-2pt}
	\item CTRL \cite{Gao_2017_ICCV}: This method considers the mean-pooled video contexts features of the moment candidate generated by a sliding window, and matches the obtained feature with the query. 
	\item SM-RL \cite{wang2019language}: This RL-based model adaptively observes the video sequence and then matches the video content with the query. 
	\item ABLR \cite{yuan2019find}: {This model incorporates a cross-modal co-attention mechanism to learn video and query attentions, which are used to localize the moment.}
	\item SAP \cite{chen2019semantic}: This method integrates the semantic concepts of the queries into the moment candidate generation to obtain discriminative candidates.	
	\item ACL \cite{ge2019mac}: This method extracts the semantic concepts from verb-obj pairs in the queries and encodes visual concepts in the video to enhance the localization.
	\item MLVI \cite{xu2019multilevel}: This multilevel language and vision integration model generates the query-specific moment candidates by incorporating the query feature to the R-C3D model \cite{xu2017r}. 
	\item TripNet \cite{hahn2019tripping}: This RL-based model utilizes the state processing module to encode the cross-modal features with gated-attention.
	\item CBP \cite{wang2020temporally}: This model predicts the boundaries based on semantic cues and aggregates contextual information through the self-attention mechanism.
	\item GDP \cite{chen2020rethinking}: The model employs a {graph convolution} to capture relationships between the multi-level semantics generated by a frame feature pyramid. 
	\item 2D-TAN \cite{zhang2020learning}: This model employs a two-dimensional temporal map to capture the temporal relations and learn more discriminative semantics of video moments
	\item SCDM \cite{yuan2019semantic}: This model employs a semantic conditioned dynamic module mechanism, which employs sentence semantics to modulate the temporal convolution process for better correlating the sentence related video contents.
	\item ExCL \cite{ghosh-etal-2019-excl}: This model predicts the frame indices of temporal boundaries from the concatenated frame features and the query feature.
	\item DRN \cite{zeng2020dense}: This method regresses the temporal distances to the boundary frames of the segment from each frame, and uses a regression model to improve the interaction between the predicted and the ground truth location.
	\item VSLNet \cite{zhang-etal-2020-span}: This method proposes a video span localizing network based on the standard span-based Question-Answering (QA) framework, and employs a query-guided highlighting strategy for prediction.
\end{itemize}

\begin{table}
	\caption{Performance comparison between the proposed model and the state-of-the-arts on TACOS dataset. The symbol '*' means the I3D features are adopted.}
	\label{table:results on tacos}
	\centering
	\begin{tabular}{c|c|c|c|c}
		\hline
		\multirow{2}{*}{Method} & R@1 & R@1 & R@1 &\multirow{2}{*}{mIoU}\\ & IoU=0.3 & IoU=0.5 & IoU=0.7 & \\
		\hline\hline
		CTRL(ICCV'2017) &18.32 &13.30 & - & -\\
		SM-RL(CVPR'2019) & 20.25 &15.95&-&-\\
		ABLR(AAAI'2019) &19.50& 9.40& - &13.40\\
		SAP(AAAI'2019) &-&18.24&-&-\\
		ACL(WACV'2019) &24.17&20.01& - &-\\
		MLVI(AAAI'2019) & 20.15 & 15.23 & - &-\\
		TripNet(CVPR'2019) & 23.95 & 19.17 & -&-\\
		CBP(AAAI'2020)&27.31&24.79&19.10& 21.59\\
		GDP(AAAI'2020)&24.14&-&-&\underline{16.18}\\
		2D-TAN(AAAI'2020)&37.29&25.32&-&-\\
		SCDM(NIPS'2019)&26.11& 21.17&-&-\\
		{ACRM\textsubscript{dt}} &{47.11}&{33.84}&{25.47}&{33.36}\\
		\hline
		ExCL*(EMNLP'2019) &\underline{45.50}&\underline{28.00}&13.80&-\\
		DRN*(CVPR'2020)&-&23.17&-&-\\	
		VSLNet*(ACL'2020)&29.61&24.27&\underline{20.03}&\underline{24.11}\\
		\hline\hline
		{ACRM\textsubscript{sg}*} &49.29&\bf{39.34}&26.12&36.44\\
		{ACRM\textsubscript{st}*} &51.09&38.37&25.82&36.59\\
		{ACRM\textsubscript{dg}*} &\bf{51.26}&38.27&26.59&37.31\\
		{ACRM\textsubscript{dt}*} &51.19&38.79&\bf{26.94}&\bf{37.42}\\
		\hline
	\end{tabular}
\end{table}
\subsubsection{Performance Analysis}
\begin{table*}
	\centering
	\caption{Ablation studies of the proposed model on Charades-STA and TACoS datasets where ATT, IFP denote the attention module and the internal-frame predictor, respectively. The ``\checkmark" symbol marks a component is enabled.}
	\label{Abalation study}
	\begin{tabular}{c|c|c|c|c|c|c||c|c|c|c}
		\hline
		\multirow{3}{*}{Method}&\multirow{3}{*}{ATT}&\multirow{3}{*}{IFP}& \multicolumn{4}{c||}{Charades-STA} & \multicolumn{4}{c}{TACoS}\\
		\cline{4-11}
		&&& R@1 & R@1 & R@1 &\multirow{2}{*}{mIoU}&R@1 & R@1 & R@1 &\multirow{2}{*}{mIoU}\\  
		&&& IoU=0.3 & IoU=0.5 & IoU=0.7 && IoU=0.3 & IoU=0.5 & IoU=0.7 &\\
		\hline\hline
		ExCL (EMNLP'2019) &&&65.10&44.10&22.60&-&45.50&28.00&13.80&-\\
		{mACRM\textsubscript{dg}} &&&71.45&55.27&35.91&51.31&48.16&37.39&24.42&35.08\\
		{mACRM\textsubscript{dt}} &&&71.96&56.16&37.63&52.13&48.46&37.44&25.52&35.33 \\
		{mACRM\textsubscript{dg} + att} &\checkmark&&71.99&56.13&36.18&51.63&49.32&38.34&26.49&36.43\\
		{mACRM\textsubscript{dt} + att} &\checkmark&&72.72&57.34&37.07&52.49&49.71&37.94&26.54&36.66\\
		{ACRM\textsubscript{dg}} &\checkmark&\checkmark&72.15&\bf{57.93}&37.15&52.60&\bf{51.26}&38.27&26.59&37.31\\
		{ACRM\textsubscript{dt}} &\checkmark&\checkmark&\bf{73.47}&57.53&\bf{38.33}&\bf{53.01}&51.19&\bf{38.79}&\bf{26.94}&\bf{37.42}\\
		\hline
	\end{tabular}
\end{table*}
The performance of four variants of our proposed model (ACRM) is reported and analyzed. {The variants are based on the different combination of interaction modeling function (i.e., element-wise multiplication or subtraction) and normalization method (i.e., $\textbf{tanh}$ or Gauss). Besides, `d', `s', `t', and `g' are used to represent element-wise multiplication, subtraction, $\textbf{tanh}$ activation, and Gauss distribution normalization, respectively. For example, ACRM\textsubscript{dg} denotes our model adopts Gauss distribution normalization and element-wise multiplication. For a fair comparison, the results of our best model ACRM\textsubscript{dt} using the C3D features are also provided.}

The results of different approaches on the Charades-STA dataset and TACoS dataset are reported in Table~\ref{table:results on Charades} and Table~\ref{table:results on tacos}, respectively. The best performance is highlighted in bold and the best results of the compared baselines are underlined. 

From the results, we can have the following observations. 
For the Charades-STA dataset, {the proposed ACRM models outperform all the competitors by a large margin in all metrics except for ACRM\textsubscript{sg}*, which fails to beat SCDM and VSLNet with a comparable result.} Compared to ExCL which uses the concatenation of the cross-modal features, the proposed models achieve around 8\%, 13\%, 16\%, and 3\% absolute improvements in terms of different metrics. {Moreover, although inferior to the results with I3D features, our C3D results still achieve significant improvements compared to all the baselines using C3D features nearly in all metrics. }

For the TACoS dataset, the four variants of the proposed model achieve the new state-of-the-art performance in terms of all metrics. Particularly, the proposed $ACRM$ model outperforms the best baseline ExCL with 6\% and 10\% improvements on the ``R@1, IoU=0.3" and ``R@1, IoU=0.5" metrics, respectively. It verifies the benefits of exploiting the cross-modal interactions and employing the internal-frame predictor to reinforce the localization process. Moreover, it is worth noting that our model makes an even larger improvement over VSLNet in the more challenging metrics ``R@1, IoU=0.7" and ``mIoU" by 6\% and 13\%, demonstrating the superiority of our model. {When using the C3D features, our model also beats all the other methods. Compared to ABLR, which uses similar attention mechanism and concatenates the attended video and query feature for boundary regression, the ACRM model achieves around 30\% improvement on the ``R@1, IoU=0.3" metric. The good performance of our model is attributed to the combining effects of the interaction modeling, the attention module for important words, and the utilization of the internal frames. In the next section, we analyze the contribution of different modules by ablation studies.}

Overall, all the variants of our method outperform the baselines consistently across all cases. {Comparing the performance of the four variants, we can see that the ACRM\textsubscript{dt}, achieves the best performance over both datasets.} Besides, using element-wise multiplication can achieve substantial improvement over the use of subtractions, indicating that the element-wise multiplication is more effective in modeling the cross-modal feature interactions.
\begin{figure}[htbp]
	\centering
	\includegraphics[width=0.99\columnwidth]{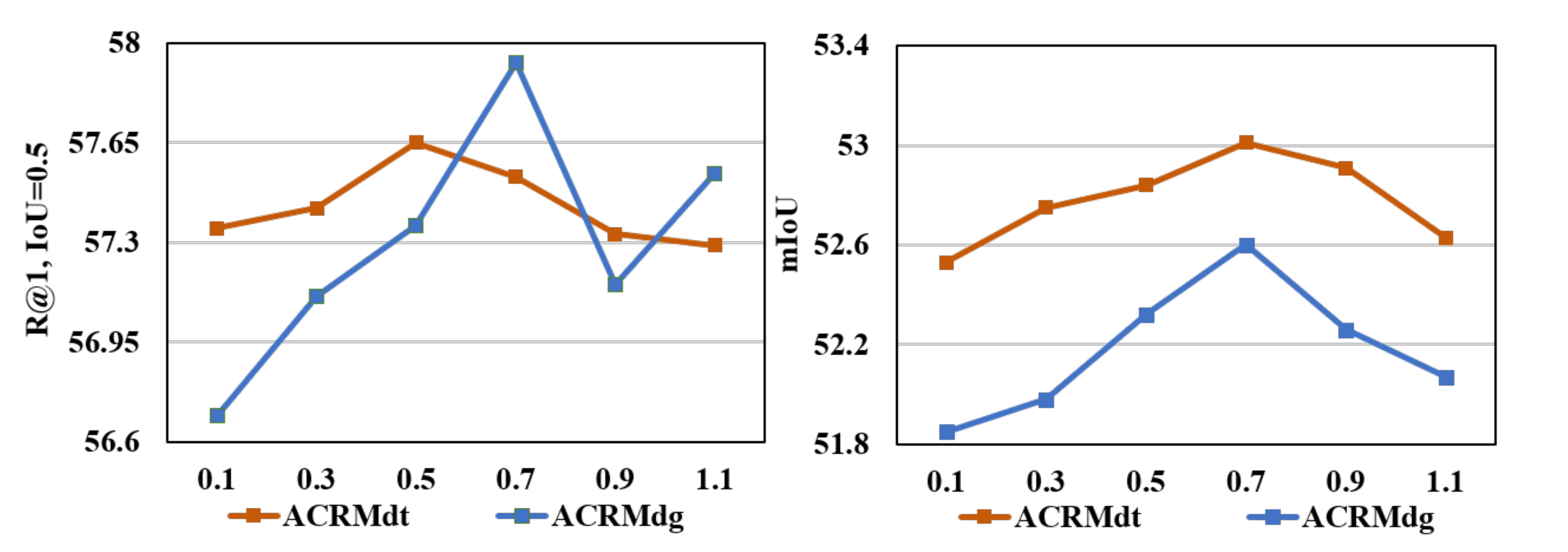}
	\caption{{The R@1, IoU=0.5 and mIoU performance of the proposed ACRM\textsubscript{dt} and ACRM\textsubscript{dg} model with the different parameter $\lambda$ on the Charades-STA dataset.}}
	\label{fig:parameteradjust}
\end{figure}
\subsection{Ablation Study}
We conduct an ablation study to examine the effectiveness of all the modules in our model, including different cross-modal interaction methods, the attention module, and the internal frame predictor. The results of the ablation study are reported in Table~\ref{Abalation study}. ExCL is used as the baseline here. For the internal-frame predictor, the influence of different trade-off hyper-parameter $\lambda$ is analyzed. We set the ExCL as the baseline where the last hidden state of the query feature and the video frame features are concatenated. Because of space limitations, {we only demonstrate the element-wise multiplication models (ACRM\textsubscript{dt} and ACRM\textsubscript{dg}), and the enabled components are marked with a ``\checkmark" symbol in Table~\ref{Abalation study}.}

%
\begin{figure*}[htbp]
	\centering
	\subfigure[The moment retrieval result of the proposed ACRM model]{
		\label{fig:visual:a} 
		\includegraphics[width=0.9\textwidth]{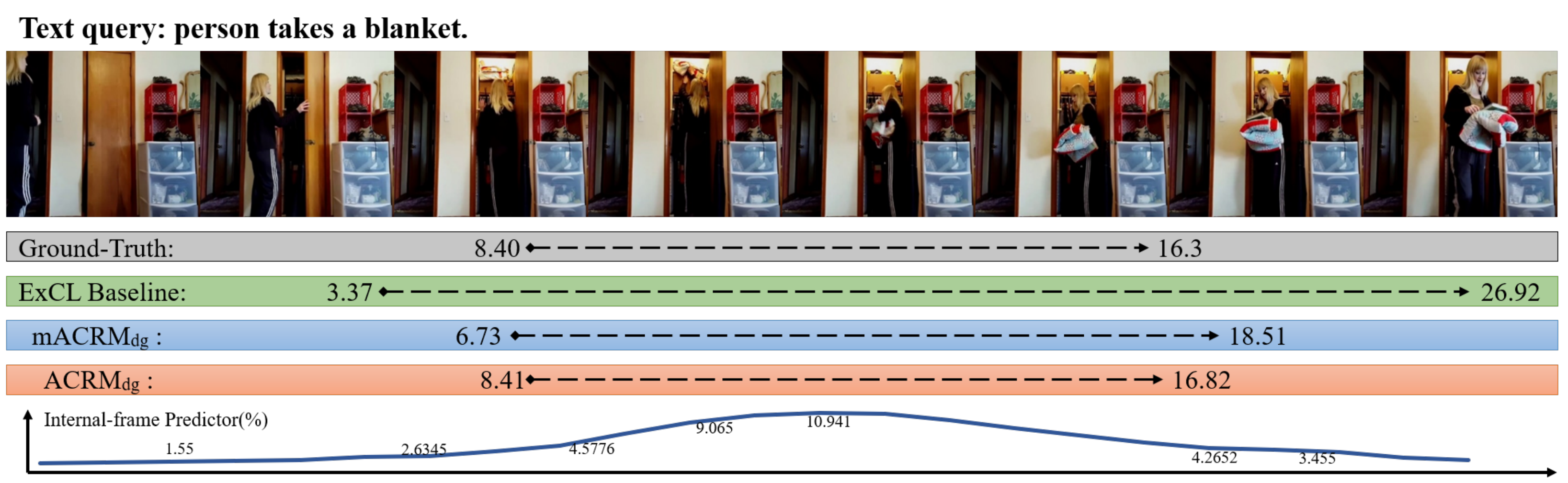}}
	\subfigure[The moment retrieval result of the proposed ACRM model]{
		\label{fig:visual:b} 
		\includegraphics[width=0.9\textwidth]{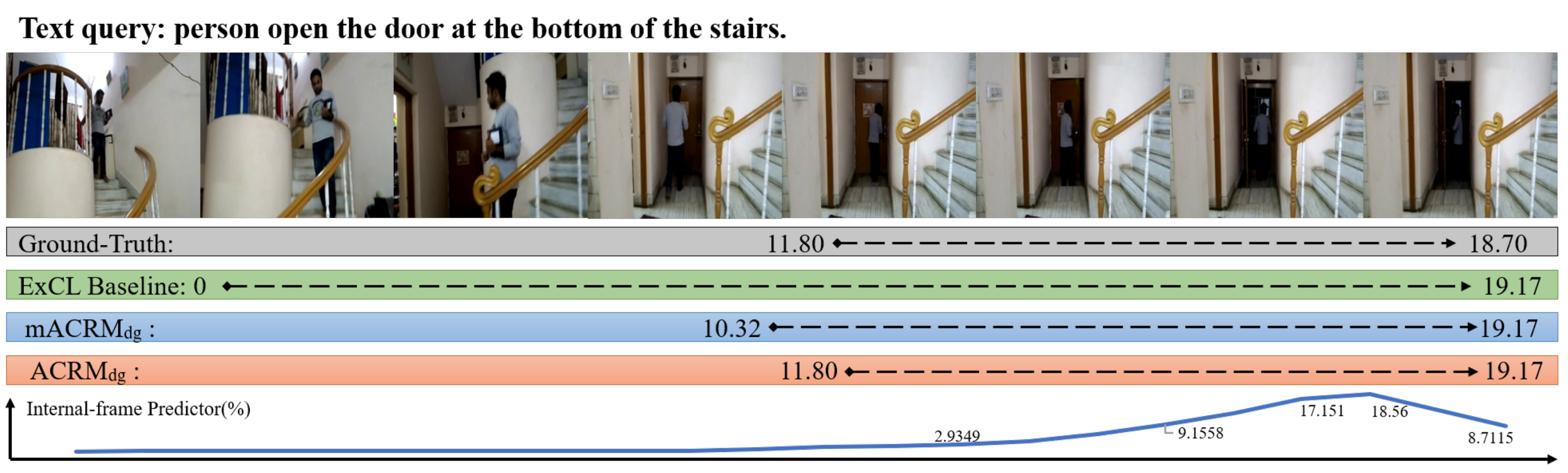}
	}
	\caption{Moment retrieval results of different models performed on the Charades-STA dataset. All the above examples are the R@1 results. The internal-frame prediction scores are obtained through ACRM\textsubscript{dg} model.}
	\label{visual}
\end{figure*}

\subsubsection{Effects of the Cross-modal Component}
{mACRM\textsubscript{dt} and mACRM\textsubscript{dg} have not used the attention module and the internal frame predictor.} Comparing to ExCL, the main difference is that ExCL uses concatenation to fuse the video frame feature and query feature, and the above two methods employ the element-wise multiplication to model the interactions between the video frame feature and mean-pooled query features. {From the comparison results on the two datasets illustrated in Table \ref{Abalation study}, we can observe that both models outperform ExCL consistently on the Charades-STA and TACoS dataset in terms of all metrics. Specifically, mACRM\textsubscript{dg} surpasses ExCL by 6.3\%, 11.1\%, 12.3\% on three metrics, respectively. Moreover, compared to all the other baselines in Table \ref{table:results on Charades} and Table \ref{table:results on tacos}, mACRM\textsubscript{dg} already achieves great improvements. It validates the importance of modeling the interactions between video features and query features in video moment retrieval instead of a simple concatenation. Particularly, the feature concatenation only maintains the unimodal cues of two modalities, which fails to obtain a stable representation. On the contrary, the proposed cross-modal interaction module can fully capture the inter-modal interactions and exploit more reliable cross-modal information than the feature concatenation.}

\subsubsection{Effects of the Attention Component}
Compared to the model without attention (i.e., the first two methods in the table), the performance can be consistently improved over all metrics when {the attention mechanism is employed (mACRM\textsubscript{dt} + att and mACRM\textsubscript{dg} + att)}. It demonstrates that the mean-pooling strategy is inadequate to exploit the correlations over the query contexts because the contributions of each word to the query representation are assumed to be equal. Therefore, the incorporation of the attention module is crucial to highlight the contribution of important words in the query and thus enhances the interactions between video frames and the corresponding query. 

\subsubsection{Effects of the Internal-frame Predictor}
The effects of the internal-frame predictor are analyzed on the Charades-STA dataset through tuning the trade-off parameter $\lambda$ in the loss function. As illustrated in Figure~\ref{fig:parameteradjust}, the performance of our ACRM model with internal-frame predictor outperforms the ones {which have not used the internal-frame predictor (mACRM\textsubscript{dt} + att and mACRM\textsubscript{dg} + att)}. This demonstrates that it is important to consider the internal frames in the modeling, which also contains useful information for boundary detection. In addition, it also indicates that although the integration of an additional internal frame predictor is simple,  it is very effective to leverage the internal frame information.

Moreover, we could also observe that when $\lambda$ increases from 0.1 to 1.1, the reported results of ``mIoU" metric only vary in the scope of 1\%, indicating the robustness of incorporate the internal-frame predictor in the model. With the increase of $\lambda$, the variation of the performance follows a general trend, i.e., rises at first and then starts to decline. The optimal value of $\lambda$ is 0.7, where all the methods obtain the best performance or the competing results on both datasets. It is expected because excessive information of the internal frames may cover the information of the boundaries frames, and thus hurt the accuracy of boundary detection. On the other side, notice that {we only incorporate the prediction score of internal frames during the training stage}. When $\lambda$ is set to a large value, the parameters of the model will be fine-tuned to predict the internal frames which will not be taken into consideration during the evaluation, resulting in performance degradation. 
\begin{figure}[htbp]
	\centering
	\includegraphics[width=0.90\columnwidth]{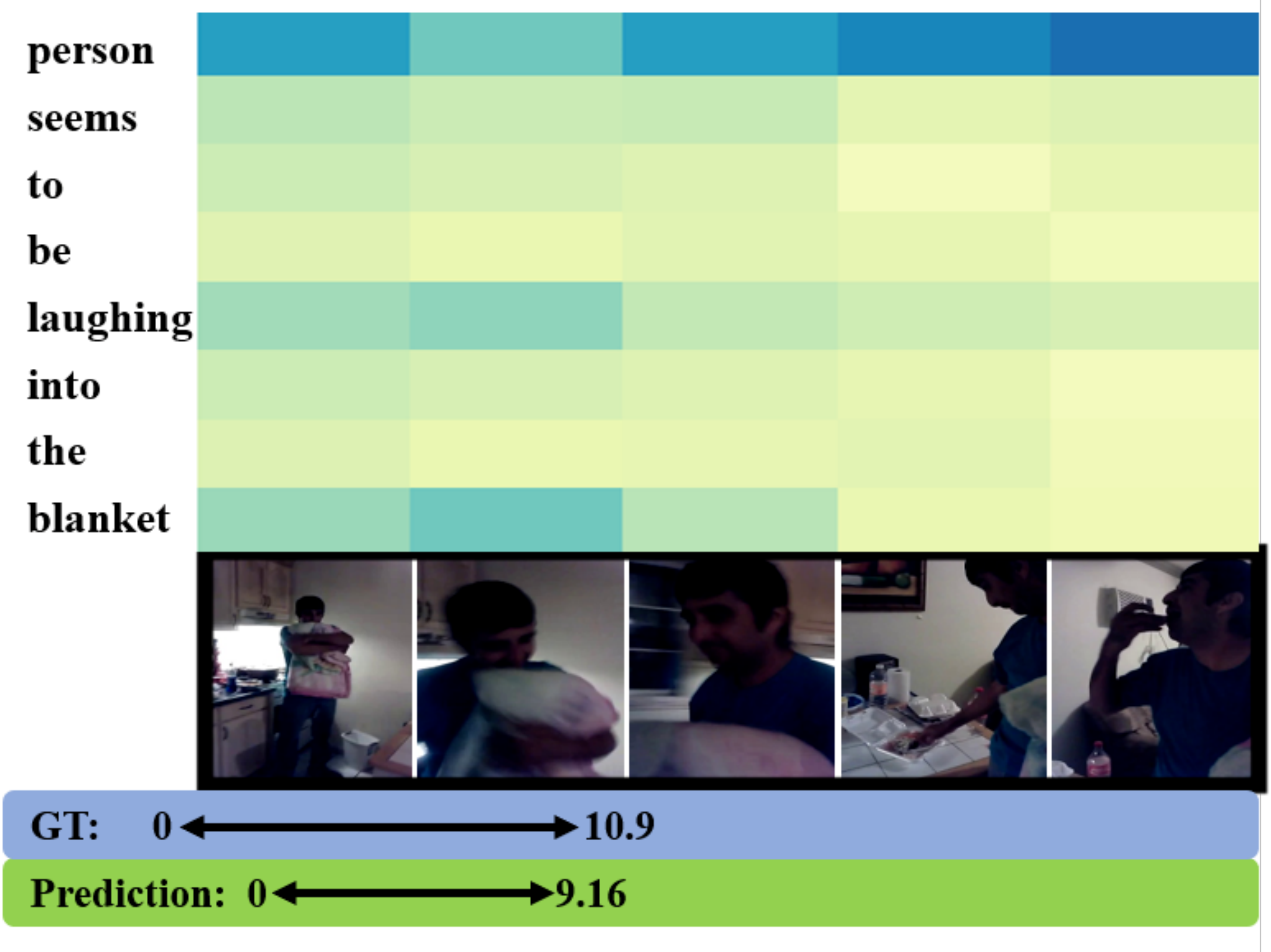}
	\caption{Visualization results of the frame-by-word attention. The darker the color is, the larger the related
		attention value is. ``GT'' represents the ground truth boundaries.}
	\label{fig:visual_att}
\end{figure}

\subsection{Qualitative Results}
\subsubsection{Visualization Results of Moment Retrieval}
In this section, some qualitative examples of the ACRM model and the baseline ExCL model for moment retrieval are illustrated in Figure~\ref{visual}. In the figure, the internal-frame prediction scores are also provided. 


Figure~\ref{fig:visual:a} presents an instance with a relatively simple query. Obviously, ExCL is incapable of returning the required moment. Instead, it returns the entire process of a woman opening her closet and leaving with a blanket because the ExCL model only maintains the information of each modality by concatenating cross-modal features, and hence fails to model the interaction between two modalities. However, the interaction is quite crucial in this video, because there are many frames with similar semantic scenes outside the desired moment, which {will confuse} the model, resulting in an inaccurate prediction of the end time. 

In contrast, {our mACRM\textsubscript{dg} model without the attention module} and the internal-frame predictor can already achieve a relatively good result with ``R@1, IoU=67.06\%", which is attributed to the use of adequate interaction modeling of two modalities. It can effectively identify the clips with high correlation and excludes those with low relevance to determine the temporal boundaries. By incorporating the attention module and the internal-frame predictor, {our full model localizes the desired moment with the highest accuracy of ``R@1, IoU=93.71\%"}. Specifically, the query attention module emphasizes the keyword ``take" which greatly enhances the prediction of the start time. As for the end time point, we also observe that the internal-frame prediction scores of the inner frames are 3-5 times higher than that of the outsiders, which indicates that the inner prediction component refines the boundaries by implicitly forcing the model to abandon the similar but irrelevant outside frames.

In addition, Figure~\ref{fig:visual:b} shows an example of a complex query. Similarly, ExCL returns the entire video, which is somehow unreasonable since the object “door” does not appear in the video at the beginning. ExCL is confused about these frames with low correlation and fails to identify the frames with higher relevance. On the contrary, {our mACRM\textsubscript{dg} model which only replaces the feature concatenation with an interaction function}, successfully excludes frames with low correlation and then {returns relatively good temporal coordinates}. It again demonstrates the importance of considering the cross-modal interactions. For our full model ACRM\textsubscript{dg}, the precise temporal boundaries are obtained, owing to the refinement of the internal-frame predictor and the attention module. 
\begin{figure}[htbp]
	\centering
	\includegraphics[width=0.90\columnwidth]{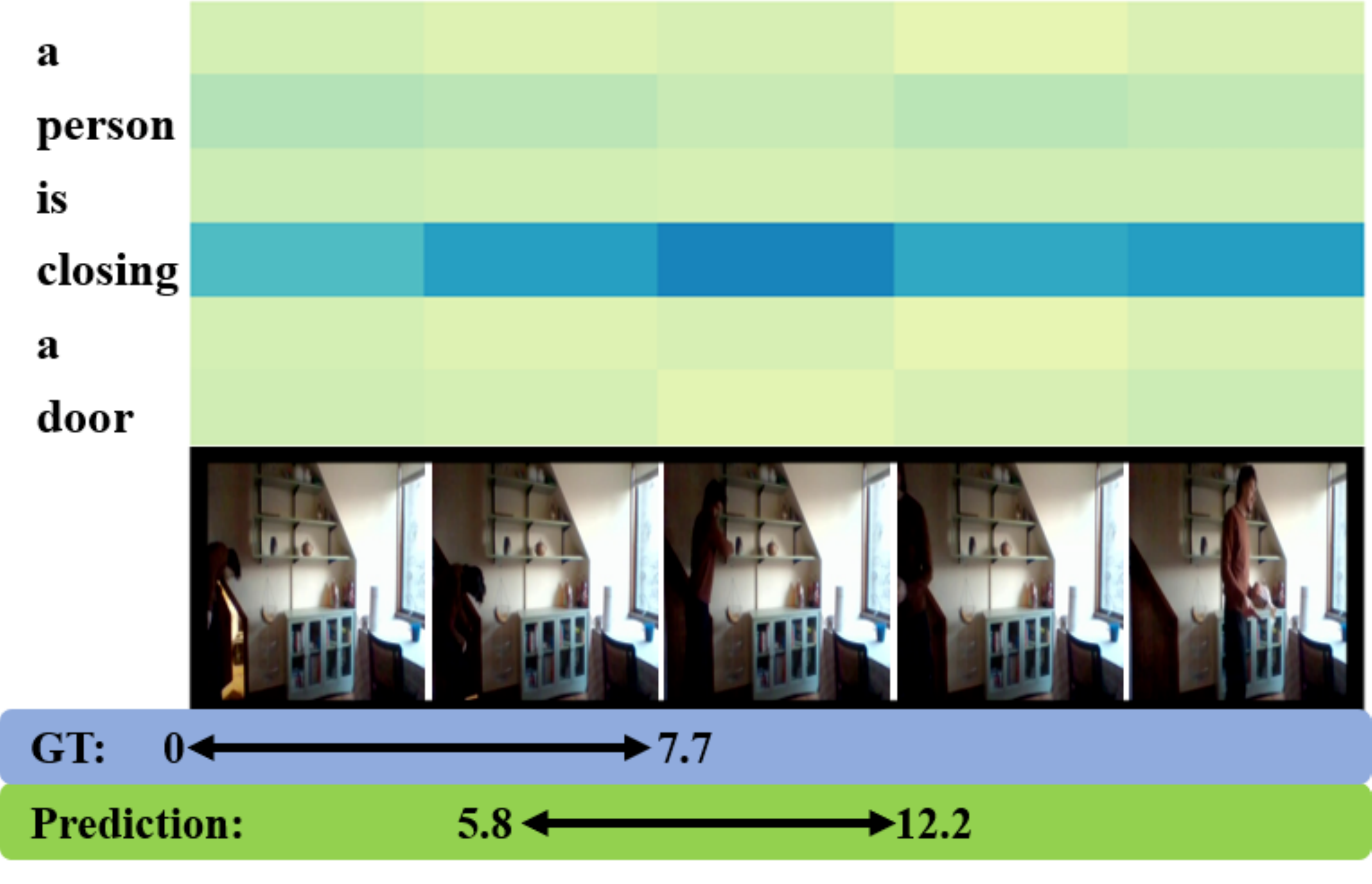}
	\caption{{Visualization results of the frame-by-word attention over a failure case. ``GT'' represents the ground truth boundaries.}}
	\label{fig:visual_att1}
\end{figure}
\subsubsection{Visualization Results of Frame-by-word Attentions}
To verify the effectiveness of the attention module, the qualitative attention weights of a video-query pair are illustrated in Figure~\ref{fig:visual_att}, where the darker the color is, the larger its represented attention weight is. As in the figure, the attention weight of the word `person' is always very large since this concept is the major object appearing across all frames. In contrast, for the words `blanket' and `laughing', our model assigns much larger attention weights in the first three frames than that in the last two frames where the concept `blanket' does not appear and the person stops `laughing'. Finally, some words like `to', `be', and `the' are very small across all frames since these words provide little information for the moment localization. {Besides, we also provide the attention weights of a failure case in Figure~\ref{fig:visual_att1}, where our method fails to capture the moment completely, and assigns high weights to the word `closing' even when the person stops the action of ``closing''. The reason might be that the visual area of the action accounts for a small proportion of the whole screen, which results in a very similar visual appearance in the entire video.} 

\section{Conclusion}
In this paper, we presented an attentive cross-modal relevance matching model (ACRM) to retrieve the relevant moment in an untrimmed video for a given specific query. Different from previous methods using a simple concatenation for boundary {prediction}, we highlight the importance of modeling the interactions between the video frame features and query features. In addition, an attention module is integrated into the model to capture more accurate frame-query relations. Moreover, our model exploits the information in the internal frame to {further enhance the model learning process} for boundary prediction by incorporating an internal-frame predictor in the objective function. Extensive experiments have been performed on two benchmark datasets to evaluate the proposed model. Experimental results show that our model substantially outperforms several state-of-the-art baselines by a large margin. Additional ablation studies also validate the effectiveness of each module in our model.

\section*{Acknowledgment}
This work is supported by the National Natural Science Foundation of China, No.: 61902223, No.: 62006142; The
Innovation Teams in Colleges and Universities in Jinan, No.:2018GXRC014; Young creative team in
universities of Shandong Province, No.2020KJN012.



\bibliographystyle{IEEEtran}
\bibliography{IEEEabrv}

%

\begin{IEEEbiography}[{\includegraphics[width=1in,height=1.25in,clip,keepaspectratio]{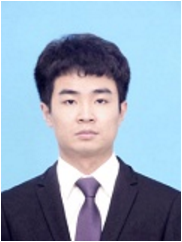}}]{Haoyu Tang} received his B.S. degree from Xi'an Jiaotong University, China, in 2016. He is currently an intern scholar in Shandong Artificial Intelligence Institute, Qilu University of Technology (Shandong Academy of Sciences), Jinan, China, and an Ph.D. candidate in software engineering department, Xi'an Jiaotong University, China. His research interests include machine learning, multimedia retrieval.
\end{IEEEbiography}

\begin{IEEEbiography}[{\includegraphics[width=1in,height=1.25in,clip,keepaspectratio]{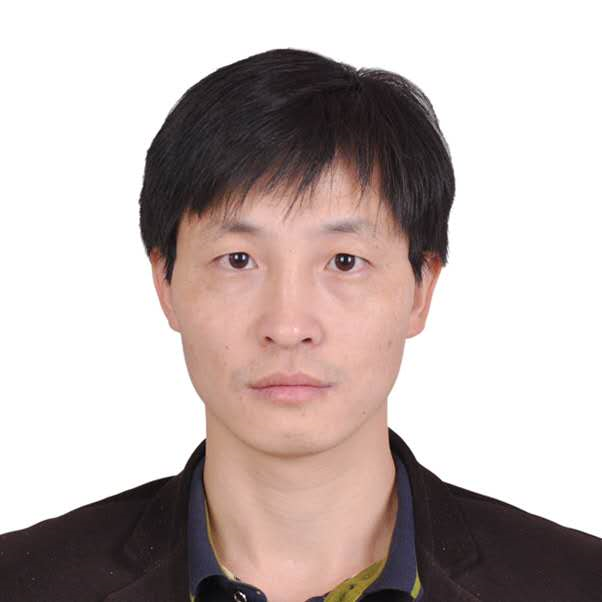}}]{Jihua Zhu}
	received the B.E. degree in automation
	from Central South University, China, and
	the Ph.D. degree in pattern recognition and intelligence system from Xi’an Jiaotong University,
	China, in 2004 and 2011, respectively. He is currently an Associate Professor with the School of
	Software Engineering, Xian Jiaotong University.
	His research interests include computer vision
	and machine learning.
\end{IEEEbiography}
\begin{IEEEbiography}[{\includegraphics[width=1in,height=1.25in,clip,keepaspectratio]{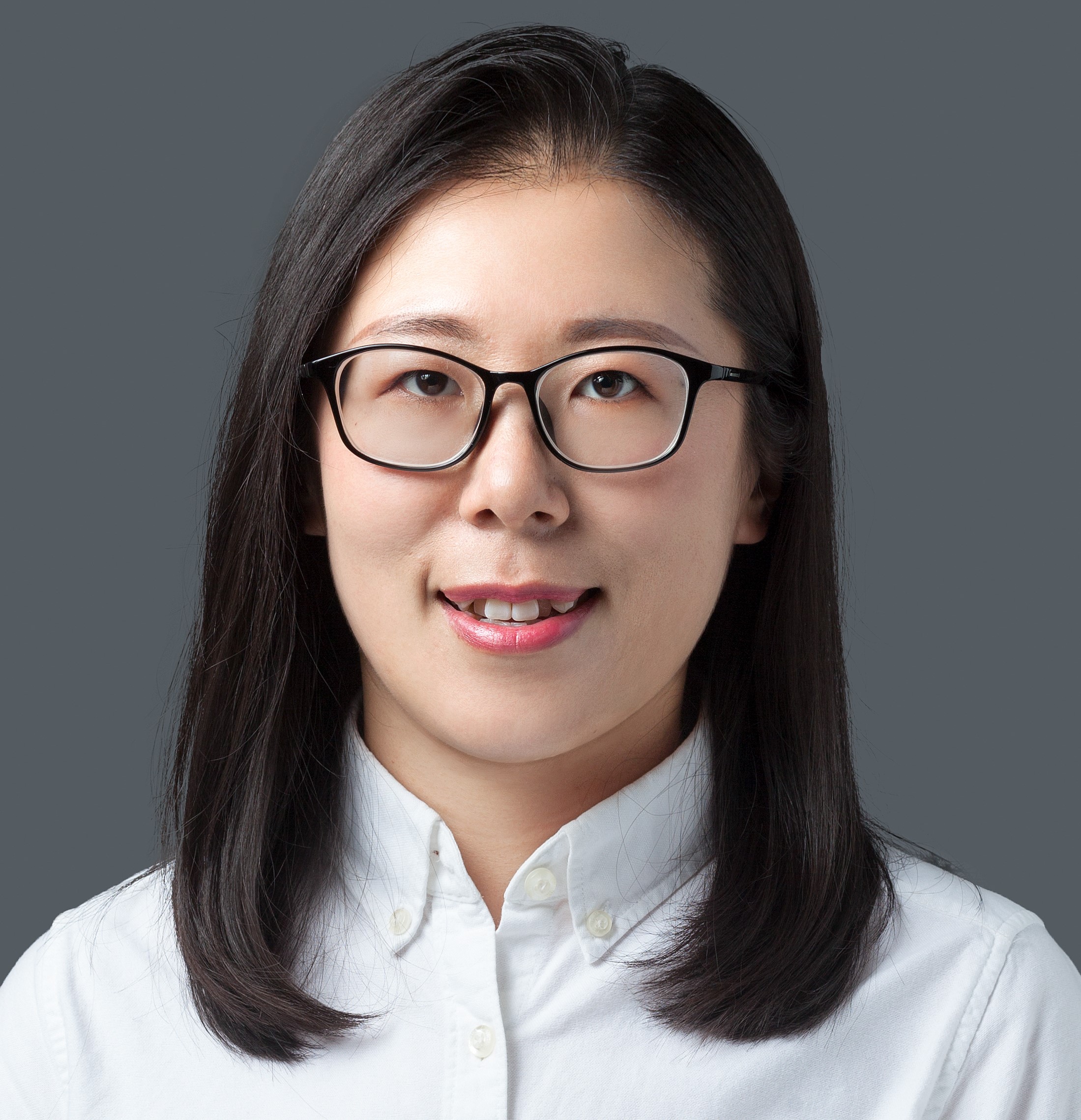}}]{Meng Liu}
 is currently a Professor with the School of Computer Science and Technology, Shandong Jianzhu University. She received the M.S. degree in computational mathematics from Dalian University of Technology, China, in 2016. Her research interests are multimedia computing and information retrieval. Various parts of her work have been published in top
forums and journals, such as SIGIR, MM, and IEEE TIP. She has served as reviewers for various conferences and journals, such as ACM MM 2019/2020, AAAI 2020, IEEE TIP, IEEE TKDE, JVCI, and INS.
\end{IEEEbiography}

\begin{IEEEbiography}[{\includegraphics[width=1in,height=1.25in,clip,keepaspectratio]{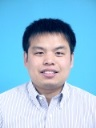}}]{Zan Gao}
 received the Ph.D. degree from the Beijing University of Posts and Telecommunications, Beijing, China, in 2011. He is currently a Full Professor with the Shandong Artificial Intelligence Institute, Shan dong Computer Science Center, Qilu University of Technology (Shandong Academy of Sciences), Jinan, China. From 2011 to 2018, he worked with the School of Computer Science and Engineering, Key Laboratory of Computer Vision and System, Ministry of Education, Tianjin University of Technology. From 2009 to 2010, he was a visiting scholar with the School of Computer Science, Carnegie Mellon University, USA, and worked with Prof. A. G. Hauptmann. From July 2016 to January 2017, he worked with Prof. T.-S. Chua with the School of Computing of National University of Singapore as a visiting scholar. His research interests include artificial intelligence, multimedia analysis and retrieval, and machine learning.
\end{IEEEbiography}

\begin{IEEEbiography}[{\includegraphics[width=1in,height=1.25in,clip,keepaspectratio]{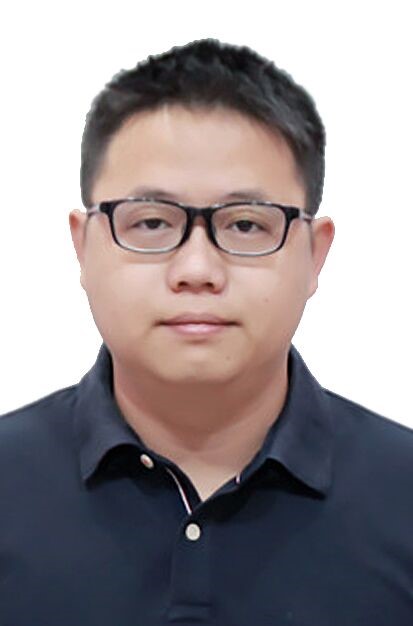}}]{Zhiyong Cheng}
	is currently a Professor with Shandong Artificial Intelligence Institute, Qilu University
	of Technology (Shandong Academy of Sciences). He
	received the Ph.D degree in computer science from
	Singapore Management University in 2016, and then
	worked as a Research Fellow in National University
	of Singapore. His research interests mainly focus on
	large-scale multimedia content analysis and retrieval.
	His work has been published in a set of top forums,
	including ACM SIGIR, MM, WWW, TOIS, IJCAI,
	TKDE, and TCYB. He has served as the PC member
	for several top conferences such as SIGIR, MM, IJCAI, AAAI. and the regular reviewer
	for journals including TKDE, TIP, TMM.
\end{IEEEbiography}

%




\end{document}